\documentclass[lettersize,journal]{IEEEtran}
\usepackage{amsmath,amsfonts}
\usepackage{algorithmic}
\usepackage{algorithm}
\usepackage{array}
\usepackage{textcomp}
\usepackage{stfloats}
\usepackage{url}
\usepackage{verbatim}
\usepackage{graphicx}
\usepackage{xcolor}
\usepackage{cite}
\usepackage{subcaption}
\usepackage{textgreek}
\usepackage{booktabs}
\usepackage{multirow}
\usepackage{array}
\newcolumntype{I}{|c|}
\usepackage{makecell}
\usepackage{graphicx}
\usepackage{subcaption}
\usepackage[utf8]{inputenc}
\usepackage[T1]{fontenc}
\usepackage[compatibility=false]{caption}

\usepackage[caption=false,font=footnotesize]{subfig}

\usepackage{tabularx}       
\usepackage{booktabs}       
\usepackage{array}          
\usepackage{ragged2e}       

\newcommand{\te}{{\rm te}}
\newcommand{\tr}{{\rm tr}}
\newcommand{\pb}{{(b)}}

\begin{document}

\title{Predictive Uncertainty in Short-Term PV Forecasting\\ under Missing Data: A Multiple Imputation Approach}

\author{Parastoo~Pashmchi,~J\'er\^ome~Benoit,~and~Motonobu~Kanagawa%
\thanks{Parastoo Pashmchi and J\'er\^ome Benoit are with SAP Labs France, Mougins, France (e-mail: pashmchi.parastoo@gmail.com, jerome.benoit@sap.com).}%
\thanks{Motonobu Kanagawa is with EURECOM, Biot, France (e-mail: motonobu.kanagawa@eurecom.fr).}%
}

\markboth{}%
{Shell \MakeLowercase{\textit{et al.}}: A Sample Article Using IEEEtran.cls for IEEE Journals}

\maketitle

\begin{abstract}
Missing values are common in photovoltaic (PV) power data, yet the uncertainty they induce is not propagated into predictive distributions. 
We develop a framework that incorporates missing-data uncertainty into short-term PV forecasting by combining stochastic multiple imputation with Rubin’s rule. 
The approach is model-agnostic and can be integrated with standard machine-learning predictors.
Empirical results show that ignoring missing-data uncertainty leads to overly narrow prediction intervals. 
Accounting for this uncertainty improves interval calibration while maintaining comparable point prediction accuracy. 
These results demonstrate the importance of propagating imputation uncertainty in data-driven PV forecasting.
\end{abstract}


\section{INTRODUCTION} \label{sec:intro}

Photovoltaic (PV) generation is a major renewable energy technology.
As its penetration increases, power system planning and control become more challenging because PV output depends on weather, location, and solar irradiance~\cite{yang2022review, antonanzas2016review}.
Accurate forecasts, together with calibrated uncertainty estimates, are therefore essential for grid operation.

Forecasting methods have evolved from persistence and classical statistical models to machine-learning approaches that capture nonlinear dynamics~\cite{akhter2019review}.
For grid integration, operational decisions (e.g., reserve scheduling) are made under forecast uncertainty and therefore require probabilistic forecasts with good calibration~\cite{wang2023quantifying, li2020review}.
Accordingly, we study probabilistic PV forecasting and prediction intervals under missing data.

\begin{figure}[t]
    \centering
    \includegraphics[width=0.9\linewidth]{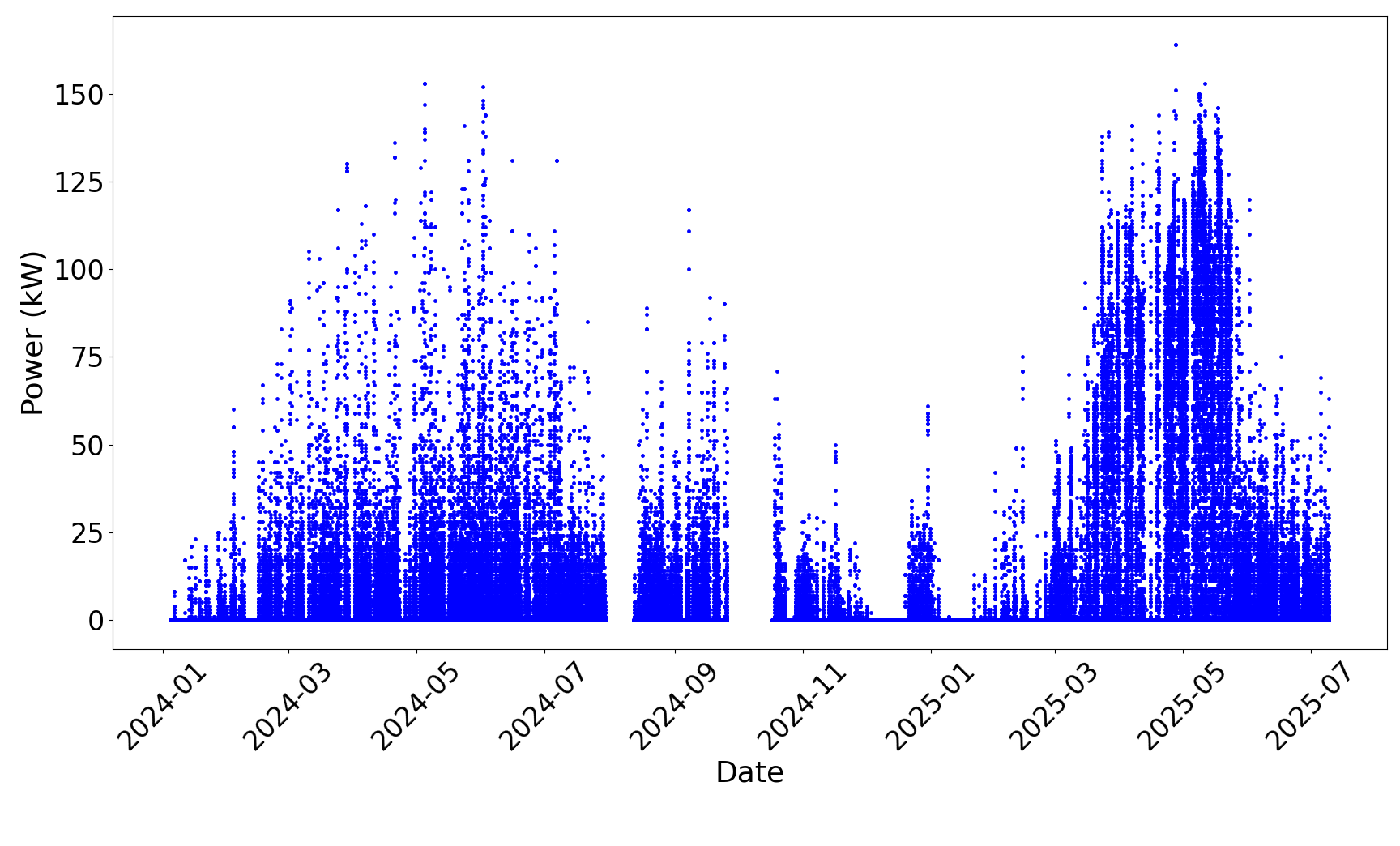}
\caption{Example of missing observations in real PV power
measurements collected at EURECOM, illustrating complete-day
gaps and prolonged zero-output periods consistent with system
outages or failures.}
    \label{fig:Eurecom_data}
\end{figure}

In practice, PV monitoring datasets often contain missing observations due to outages, equipment faults, and communication or logging failures~\cite{koumpli2016missing,livera2021dataquality, IEA_PVPS_T13_22_2021,lin2025dataset}. 
Missingness can exceed 10\%~\cite{IEA_PVPS_T13_22_2021,livera2021dataquality} and severe cases considered in the literature reach 40\%~\cite{livera2021dataquality};  gaps may last weeks to months~\cite{koumpli2016missing}.
Figure~\ref{fig:Eurecom_data} illustrates such patterns in real PV measurements, including complete-day gaps and prolonged zero-output periods.

Missing values are not just a preprocessing issue; they are a source of predictive uncertainty.
In PV forecasting, missingness can affect both training data and test-time inputs.
Single imputation (e.g., zeros or averages) treats unknown values as fixed, typically understating predictive variance and yielding miscalibrated prediction intervals.

A substantial body of work develops probabilistic PV forecasting under complete-data assumptions.
For example, existing approaches based on Gaussian process models \cite{najibi2021enhanced}, ensemble-based approaches \cite{mayer2022probabilistic}, spatio-temporal probabilistic models \cite{agoua2018probabilistic}, 
interval forecasting methods including bootstrap- and quantile-regression-based constructions \cite{han2019pv,wen2019performance}, 
nonparametric density estimation \cite{golestaneh2016very}, and uncertainty-aware day-ahead and short-term models \cite{gu2021forecasting,liu2018prediction} all quantify predictive uncertainty.
However, these approaches assume fully observed data
and do not account for uncertainty induced by missing observations.

\subsection{Existing Works on Missing Values in PV Systems}
A separate line of research addresses missing data in PV systems~\cite{zhang2020solargan,shen2021missing,benitez2023novel, shireen2018iterative, lee2024pv, liu2022missing, liu2021pv,costa2024employing,phan2024enhancing}. 
Imputation is performed prior to forecasting, after which the completed dataset is treated as fully observed. Predictive performance is evaluated by point-error metrics. Uncertainty arising from the imputation step is neither quantified nor incorporated into the forecasting model.

Zhang et al.~\cite{zhang2020solargan} propose SolarGAN, a WGAN-based imputation method for multivariate solar time series.
A recurrent generator conditions on the observed entries (with added noise) to fill missing values, while a discriminator enforces realistic joint dynamics across variables, yielding a single completed dataset.
Imputation uncertainty is not quantified or propagated into prediction intervals.

Shen et al.~\cite{shen2021missing} propose a deep generative reconstruction model that imputes missing values in test-time inputs prior to forecasting by conditioning on the observed variables within a multi-modal time window, including PV output, meteorological measurements (such as solar radiation), and sky images, without quantifying imputation uncertainty.

Benitez et al.~\cite{benitez2023novel} use irradiance and weather variables to fill missing PV outputs by averaging historical observations under similar conditions, and then forecast PV output from the resulting completed dataset using a point-prediction model; no uncertainty quantification is introduced for either the imputation or the forecasting stage.

Shireen et al.~\cite{shireen2018iterative} use an iterative multi-task Gaussian-process framework to estimate missing historical PV observations from correlated panels and irradiance data, and then apply ARIMA to forecast future PV output based on the imputed series; no uncertainty from the imputation step is propagated into the forecasting model.

Lee et al.~\cite{lee2024pv} study missing PV power data by imputing it with deletion, linear interpolation, kNN, or GAIN using irradiance and temperature measurements and weather-forecast variables (including precipitation and sky status), and then train a CNN--GRU point-forecasting model on the completed data; no uncertainty quantification is provided for either imputation or forecasting.

Liu et al.~\cite{liu2022missing, liu2021pv} address missing values in short PV output histories used as inputs for forecasting.
A convolutional neural network is trained on complete data with artificially introduced gaps and then used to reconstruct missing entries in the input series.
Future PV output is predicted from the reconstructed data using a regression model.
Both reconstruction and forecasting are formulated as point estimation, and no predictive uncertainty is quantified.

Costa et al.~\cite{costa2024employing} predict missing PV generation values from meteorological variables using tree-based regressors trained on complete data.
The approach yields a single imputed value for each gap and does not quantify imputation uncertainty.

Phan et al.~\cite{phan2024enhancing} handle missing values in PV and meteorological input series by an iterative MICE-style procedure.
Missing entries are first initialized by mean imputation, after which the completed dataset is repeatedly resampled by bootstrap; at each step an XGBoost regressor is fitted and predictive mean matching is used to update the imputed values.
The resulting data are treated as a single completed dataset for downstream forecasting.
Prediction intervals are produced by a Transformer–LUBE model.
Bootstrap is used to repeat the imputation procedure rather than to quantify or propagate uncertainty in the imputed values.

\subsection{Contributions}

\begin{figure}[t]
    \centering
    \includegraphics[width=0.95\linewidth]{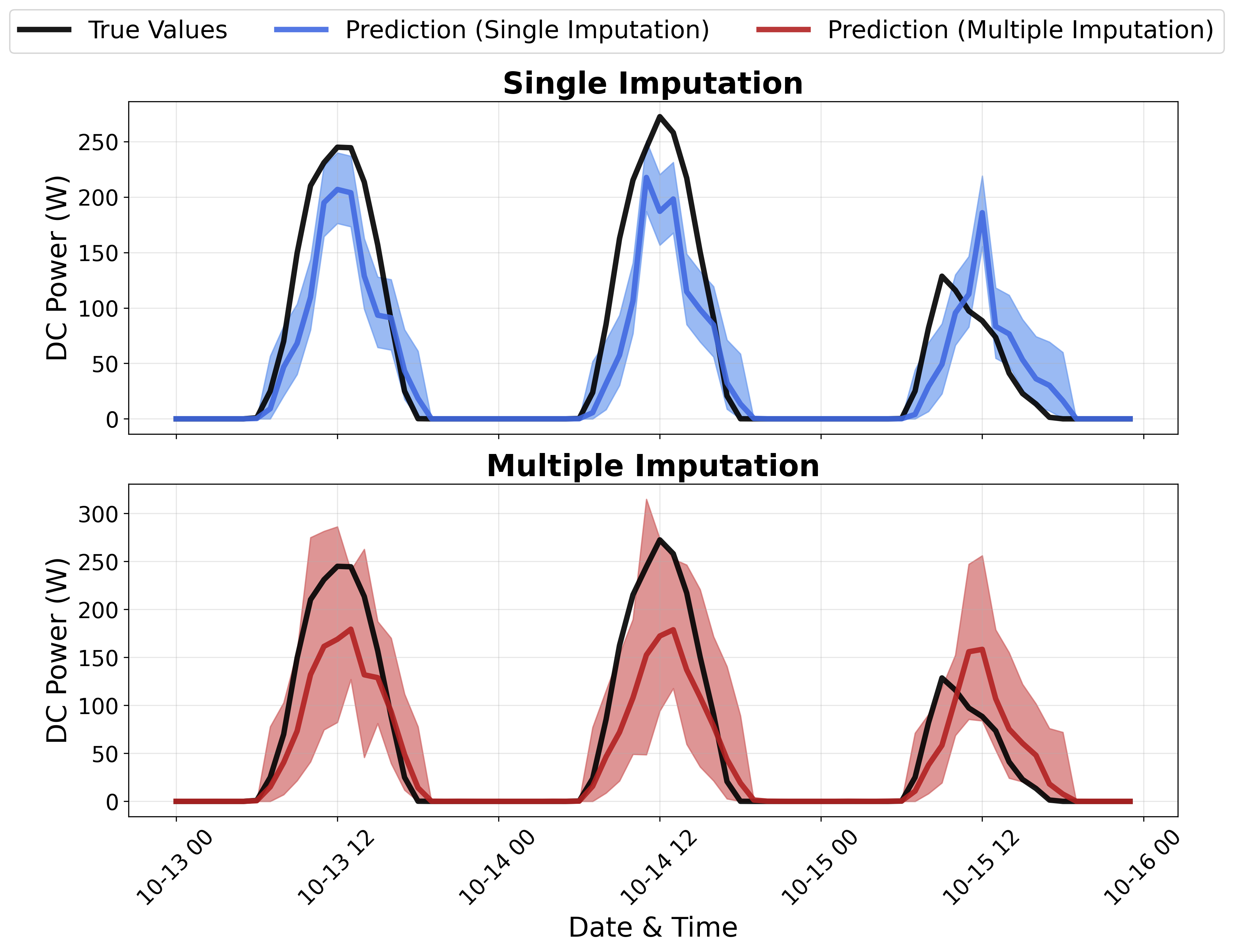}
\caption{Comparison of single and multiple imputation for one-hour-ahead prediction with a Random Forest model. The shaded regions show the 95\% prediction intervals. Single imputation gives overly narrow intervals, whereas the proposed multiple-imputation approach accounts for missing-value uncertainty and gives wider intervals.}
    \label{fig:intro-RF-single-vs-multiple-imputation}
\end{figure}

While missing values are recognized as a practical issue in PV forecasting, the uncertainty they induce is neither quantified nor propagated into predictive distributions.
Missing values may arise in both the training data—including inputs and outputs—and in test-time inputs. In such cases, these variables should be regarded as uncertain rather than fixed.
Replacing missing entries with single imputed values underestimates predictive variance, resulting in overly narrow prediction intervals. Given the substantial variability of PV generation, poorly calibrated intervals may distort grid operation, for example through insufficient reserve allocation.

This study develops a framework for propagating missing-data uncertainty into predictive distributions in short-term PV forecasting. 
Multiple imputation is widely used for parameter inference under missing data~\cite{rubin1987multiple}. 
Here we adapt this principle to predictive uncertainty in a model-agnostic manner.
Figure~\ref{fig:intro-RF-single-vs-multiple-imputation} illustrates the main effect of the proposed multiple-imputation approach: single imputation yields overly narrow predictive intervals, whereas multiple imputation accounts for missing-value uncertainty and gives wider intervals. Experimental details are given in Section~\ref{sec:experiment-setup}.

The main contributions are as follows:

\begin{itemize}

\item \textbf{Principled propagation of missing-data uncertainty.} 
We combine stochastic multiple imputation with Rubin’s rule to incorporate uncertainty due to missing inputs and targets into predictive variance.

\item \textbf{Unified and model-agnostic formulation.}
The framework handles missing values in both training (inputs and targets) and test-time inputs, and can be integrated with standard machine-learning predictors.

\item \textbf{Implications for predictive calibration.}
Ignoring missing-data uncertainty results in overly narrow intervals. 
Accounting for this uncertainty improves calibration without degrading point prediction accuracy.

\end{itemize}

The remainder of this paper is organized as follows. 
Section~\ref{sec:setup-prediction-complete-data} presents the one-hour-ahead PV forecasting setup under complete data and establishes the notation. 
Section~\ref{sec:methodology} develops the multiple imputation framework for predictive uncertainty quantification in the presence of missing values. 
Section~\ref{sec:experiment-setup} outlines the experimental design, and Section~\ref{sec:results} reports the empirical findings.

\section{Forecast Setup without Missing Values}
\label{sec:setup-prediction-complete-data}

We first describe the PV forecasting task under complete data and introduce the notation. Missing values are incorporated in the next section.

\subsection{One-hour-ahead PV Power Forecasting}

We consider one-hour-ahead PV power forecasting using the previous 24 hours of PV power and irradiance. Thus, the response is the PV power in the next hour, and the input is a 48-dimensional vector consisting of 24 hourly PV power values and 24 hourly irradiance values. Each observation is the total PV energy or irradiance over the corresponding hour.

\subsection{Machine Learning Training} 
\label{sec:training-complete}

Let
\begin{equation} \label{eq:data-train-period-complete}
     (P_1, I_1), (P_2, I_2), \dots, (P_T, I_T)
\end{equation}
be a historical time series of hourly PV power and irradiance, where $P_t \geq 0$ and $I_t \geq 0$ denote the total PV power and irradiance at hour $t$, respectively, for $t = 1, \dots, T$.
From this series, we construct input-output pairs for one-hour-ahead supervised learning: the input is the previous 24 hours of PV power and irradiance, and the output is the PV power in the next hour.

We define the training pairs by
\begin{align} 
    X_t &= ( P_t, I_t, P_{t-1}, I_{t-1}, \dots, P_{t-23}, I_{t-23} )^\top \in \mathbb{R}^{48}, \nonumber \\
    Y_t &= P_{t+1} \in \mathbb{R}, \label{eq:train-pairs-comp} 
\end{align}
for $t = 24, \dots, T-1$.
The training dataset is
\[
D_{\tr} := \{ (X_{24},Y_{24}), \dots, (X_{T-1}, Y_{T-1}) \},
\]
with $N = T - 24$ pairs.

A prediction model $\hat f$ is trained on $D_{\tr}$ to estimate the relation between $X_t$ and $Y_t$. In general, $\hat f$ is obtained by minimizing a training loss over a model class $\mathcal{F}$, for example,
\begin{equation} \label{eq:emp-risk-min}
    \hat{f} = \arg\min_{f \in \mathcal{F}} \frac{1}{N} \sum_{i=1}^N \left( f(X_i) - Y_i \right)^2,
\end{equation}
when squared-error loss is used.

We write the trained model as
\begin{equation} \label{eq:training-operator}
    \hat f = {\rm Train}_{\mathcal F}(D_{\tr}),
\end{equation}
where ${\rm Train}_{\mathcal F}$ denotes the training procedure applied to $D_{\tr}$.

\subsection{Machine Learning Forecast on Test Data}  
\label{sec:ML-pred-test-comp}

Let
\begin{align} \label{eq:test-data-complete}
    (P^{\te}_1, I^{\te}_1), (P^{\te}_2, I^{\te}_2), \dots, (P^{\te}_{T'}, I^{\te}_{T'})
\end{align}
be the test-period time series of hourly PV power and irradiance.
For $t=24,\dots,T'-1$, the next-hour PV power
\[
Y_t^{\te}=P_{t+1}^{\te}
\]
is predicted from
\begin{align} \label{eq:ML-pred-comp}
X_t^{\te} = ( P_{t}^{\te}, I_{t}^{\te}, P_{t-1}^{\te}, I_{t-1}^{\te}, \dots, P_{t-23}^{\te}, I_{t-23}^{\te} )^\top,
\end{align}
that is,
\[
\hat f(X_t^{\te}) \approx Y_t^{\te}.
\]

To quantify predictive uncertainty, we use a Gaussian predictive distribution with mean $\hat f(X_t^{\te})$ and variance estimated from the training residuals:
\begin{equation} \label{eq:train-res-var-comp}
    \hat{\sigma}^2 := \frac{1}{N} \sum_{i=1}^N \left( \hat{f}(X_i) - Y_i \right)^2 .
\end{equation}
This is the maximum likelihood estimator under a Gaussian noise model. Other probabilistic forecasting approaches are also available; see, for example, \cite{najibi2021enhanced,mayer2022probabilistic,agoua2018probabilistic,han2019pv,wen2019performance,golestaneh2016very,gu2021forecasting,liu2018prediction}.

\section{Multiple Imputation for PV Forecasting}
\label{sec:methodology}

In practice, many PV power values are missing, so the procedures in Section~\ref{sec:setup-prediction-complete-data} are not directly applicable. Removing incomplete observations reduces the training sample size and may prevent test-time forecasting. We therefore impute the missing values and quantify the resulting uncertainty. This section describes our multiple imputation approach.

\subsection{Stochastic Imputation of Missing PV Power Values} \label{sec:imputation-knnsampler}

PV power values may be missing in both training and test data. We represent missingness by binary indicators:
\begin{align}
& \text{Training:}~  (P_1, I_1, M_1), (P_2, I_2, M_2), \dots, (P_T, I_T, M_T),  \label{eq:train-miss} \\
& \text{Test:}~   (P^{\te}_1, I^{\te}_1, M^{\te}_1), (P^{\te}_2, I^{\te}_2, M^{\te}_2), \dots, (P^{\te}_{T'}, I^{\te}_{T'}, M^{\te}_{T'}), \label{eq:test-missing}
\end{align}
where $M_t=1$ if $P_t$ is missing and $M_t=0$ otherwise; similarly, $M_t^{\te}=1$ if $P_t^{\te}$ is missing and $M_t^{\te}=0$ otherwise.

Missing PV power values are imputed from the statistical relation between PV power and irradiance in the training data. A natural target is the conditional distribution of PV power given irradiance:
\[
\hat P_t \sim \Pr(P_t \mid I_t), \qquad t=1,\dots,T \ \text{with}\ M_t=1.
\]
Since this conditional distribution is unknown, it must be estimated from the observed data. In our experiments, we use kNNSampler~\cite{pashmchi2025knnsampler}, a stochastic imputation method with theoretical convergence guarantees. Other consistent estimators could also be used, and the conditioning variables could be extended beyond contemporaneous irradiance.

kNNSampler imputes a missing PV power value at irradiance $I_t$ by sampling from the observed PV power values corresponding to the $k$ nearest irradiance neighbors of $I_t$. Equivalently, it approximates the conditional distribution by
\begin{equation} \label{eq:kNN-cond}
{\rm Pr}(P_t \mid I_t) \approx \widehat{\rm Pr}(P_t \mid I_t)
= \frac{1}{k} \sum_{i \in {\rm NN}(I_t, k)} \delta_{P_i},
\end{equation}
where $\delta_{P_i}$ is the point mass at $P_i$, and ${\rm NN}(I_t, k)$ is the set of indices of the $k$ nearest irradiance observations to $I_t$ among those with observed PV power.

Under mild regularity conditions, the kNNSampler empirical distribution converges to the true conditional distribution as $n \to \infty$, $k \to \infty$, and $k/n \to 0$, where $n$ is the number of observed irradiance--PV power pairs. In practice, $k$ is a hyperparameter selected by the fast leave-one-out cross-validation method of \cite{kanagawa2024fast}; see \cite{pashmchi2025knnsampler} for details.

\subsection{Multiple Imputation Framework}
\label{sec:multiple-imput-frame}

We now describe the proposed multiple imputation framework. It generates $B \geq 1$ completed versions of the training and test data, leading to $B$ predictions and uncertainty estimates for each test instance. These are combined into a single prediction and uncertainty estimate that reflects missing-value uncertainty. The case $B=1$ corresponds to single imputation.

In what follows, $b=1,\dots,B$ indexes the imputation rounds.

\subsubsection{Imputing Missing Training Data}

For each imputation round $b=1,\dots,B$, missing PV power values in the training data~\eqref{eq:train-miss} are imputed by sampling from the estimated conditional distribution given the observed irradiance:
\[
\hat{P}_t^{(b)} \sim \widehat{\rm Pr}(P_t \mid I_t),
\qquad t=1,\dots,T \ \text{with}\ M_t=1.
\]
Define
\[
P_s^{(b)} =
\begin{cases}
P_s, & \text{if } M_s=0,\\
\hat P_s^{(b)}, & \text{if } M_s=1.
\end{cases}
\]
The $b$-th completed training dataset is then constructed as
\[
D_{\tr}^{(b)} := \{(X_{24}^{(b)},Y_{24}^{(b)}),\dots,(X_{T-1}^{(b)},Y_{T-1}^{(b)})\},
\]
where, for $t=24,\dots,T-1$,
\begin{align*}
X_t^{(b)} &= (P_t^{(b)}, I_t, P_{t-1}^{(b)}, I_{t-1}, \dots, P_{t-23}^{(b)}, I_{t-23})^\top,\\
Y_t^{(b)} &= P_{t+1}^{(b)}.
\end{align*}

A prediction model is trained on the $b$-th completed training dataset, which we write as
\[
\hat f^{(b)} = {\rm Train}_{\mathcal F}(D_{\tr}^{(b)}).
\]

\subsubsection{Imputing Missing Test Input Features}

At test time, some of the past 24 PV power values used as input may be missing. For each imputation round $b=1,\dots,B$, we define  
\[
P_s^{\te (b)} =
\begin{cases}
\hat P_s^{\te (b)} \sim \widehat{\rm Pr}(P_s^{\te}\mid I_s^{\te}), & \text{if } M_s^{\te}=1,\\
P_s^{\te}, & \text{if } M_s^{\te}=0,
\end{cases}
\]
for each $s=t,t-1,\dots,t-23$.
The resulting completed test input is
\begin{align} \label{eq:pred-test-imp}
X_t^{\te (b)} = (P_t^{\te (b)}, I_t^{\te}, P_{t-1}^{\te (b)}, I_{t-1}^{\te}, \dots, P_{t-23}^{\te (b)}, I_{t-23}^{\te})^\top,
\end{align}
and the next-hour PV power is predicted by
\[
\hat f^{(b)}(X_t^{\te (b)}) \approx Y_t^{\te} = P_{t+1}^{\te}.
\]

For each imputation round, predictive uncertainty is quantified by a variance estimate. Here we use the training residual variance computed on the $b$-th completed training dataset:
\begin{equation} \label{eq:train-res-var-imp}
    (\hat{\sigma}^2)^{\te \pb}
    =
    \frac{1}{N} \sum_{i=1}^N \left( \hat f^\pb(X_i^\pb) - Y_i^\pb \right)^2 .
\end{equation}
This is the same residual-based variance estimate as in \eqref{eq:train-res-var-comp}, now applied to the imputed training data. More sophisticated probabilistic forecasting methods could also be used here; see, for example, \cite{najibi2021enhanced,mayer2022probabilistic,agoua2018probabilistic,han2019pv,wen2019performance,golestaneh2016very,gu2021forecasting,liu2018prediction}.

\subsection{Aggregation by Rubin's Rule}

For each imputation round $b=1,\dots,B$, we obtain a predictive mean and variance. These are then combined by Rubin's rule~\cite{rubin1987multiple} to yield the final predictive mean and variance.

The final predictive mean is the average of the $B$ predictive means:
\begin{equation} \label{eq:pred-mean-final}
    \hat{Y}_t^\te = \frac{1}{B} \sum_{b=1}^B \hat{f}^{\pb}(X_t^{\te \pb}) .
\end{equation}

The final predictive variance combines the within-imputation variance (WV) and the between-imputation variance (BV). The within-imputation variance is
\[
\widehat{\rm WV}_t^\te = \frac{1}{B} \sum_{b=1}^B (\hat{\sigma}^2)^{\te \pb},
\]
the average of the $B$ predictive variances. The between-imputation variance is
\[
\widehat{\rm BV}_t^\te = \frac{1}{B-1} \sum_{b=1}^B \left( \hat{Y}_t^\te - \hat{f}^{\pb}(X_t^{\te \pb}) \right)^2,
\]
the sample variance of the $B$ predictive means. Rubin's rule then defines the total predictive variance by
\begin{equation} \label{eq:pred-variance-final}
(\hat{\sigma}_t^2)^\te = \widehat{\rm WV}_t^\te + \left(1 + \frac{1}{B}\right)\widehat{\rm BV}_t^\te .
\end{equation}

Rubin's rule~\eqref{eq:pred-variance-final} is a finite-sample analogue of the law of total variance~\cite{rubin1987multiple,Rubin96}. Let $Z_t$ denote the missing values that affect prediction at time $t$. Then, conditional on the observed data,
\[
\operatorname{Var}(Y_t^\te)
=
\mathbb{E}\!\left[ \operatorname{Var}(Y_t^\te \mid Z_t) \right]
+
\operatorname{Var}\!\left( \mathbb{E}[Y_t^\te \mid Z_t] \right).
\]
The first term corresponds to the within-imputation variance, and the second to the between-imputation variance. Rubin's rule estimates this decomposition with a finite number of imputations; the factor $1+\frac{1}{B}$ in~\eqref{eq:pred-variance-final} is the corresponding finite-$B$ correction.

Single imputation is the special case in which there is no between-imputation variability. Then
\[
\operatorname{Var}\!\left( \mathbb{E}[Y_t^\te \mid Z_t] \right)=0,
\]
and Rubin's rule reduces to
\[
(\hat{\sigma}_t^2)^\te = \widehat{\rm WV}_t^\te .
\]
Thus, only within-imputation uncertainty is retained.

\label{methodology:uncertainty}

\subsection{Predictive Intervals}\label{sec:interval_estimation}

A predictive interval for next-hour PV power is obtained from a distribution whose mean and variance match the final predictive mean~\eqref{eq:pred-mean-final} and variance~\eqref{eq:pred-variance-final}. We consider normal and gamma distributions. The normal distribution is widely used and gives better-calibrated intervals in our experiments, but it may produce negative lower bounds. The gamma distribution is included as a simple non-negative baseline, since it is supported on non-negative values.

We derive a $100(1-\alpha)\%$ predictive interval for $0<\alpha<1$, for example the $95\%$ interval when $\alpha=0.05$.

\subsubsection{Normal-based Intervals}

Under a normal distribution with predictive mean~\eqref{eq:pred-mean-final} and variance~\eqref{eq:pred-variance-final}, the $100(1-\alpha)\%$ predictive interval is
\[
\left[
\hat{Y}_t^\te - z_{1-\alpha/2}\sqrt{(\hat{\sigma}_t^2)^\te},
\quad
\hat{Y}_t^\te + z_{1-\alpha/2}\sqrt{(\hat{\sigma}_t^2)^\te}
\right],
\]
where $z_{1-\alpha/2}$ is the $(1-\alpha/2)$ quantile of the standard normal distribution (e.g., $z_{0.975}\approx 1.96$).

Normal-based intervals are widely used and, in our experiments, better calibrated in terms of coverage. However, they are symmetric about the mean and may include negative values. Since PV power is non-negative and typically asymmetric, we also consider a gamma-based interval.

\subsubsection{Gamma-based Intervals}

For $y \ge 0$, let
\[
f(y; a_t,b_t)
=
\frac{y^{a_t-1}e^{-y/b_t}}{\Gamma(a_t)b_t^{a_t}},
\qquad a_t>0,\ b_t>0,
\]
be the gamma density with shape $a_t$ and scale $b_t$; see, for example, \cite{krishnamoorthy2006handbook}. Its mean and variance are $a_t b_t$ and $a_t b_t^2$, respectively. Matching these to the predictive mean~\eqref{eq:pred-mean-final} and variance~\eqref{eq:pred-variance-final} gives
\[
a_t = \frac{(\hat{Y}_t^\te)^2}{(\hat{\sigma}_t^2)^\te},
\qquad
b_t = \frac{(\hat{\sigma}_t^2)^\te}{\hat{Y}_t^\te}.
\]
With these parameters, the $100(1-\alpha)\%$ predictive interval is
\[
\Bigl[
G^{-1}\bigl(\tfrac{\alpha}{2};\,a_t,b_t\bigr),
\;G^{-1}\bigl(1-\tfrac{\alpha}{2};\,a_t,b_t\bigr)
\Bigr],
\]
where $G^{-1}(p;a_t,b_t)$ denotes the $p$th quantile of the gamma distribution with parameters $a_t$ and $b_t$. If the predictive mean $\hat{Y}_t^\te$ is non-positive, we use the degenerate interval $[0,0]$.

\section{Experiment Setup}

\label{sec:experiment-setup}

\subsection{Dataset Implementation} {\label{sec:data-serbia}}

\begin{figure}[t]
    \centering
        \includegraphics[width=1\linewidth]{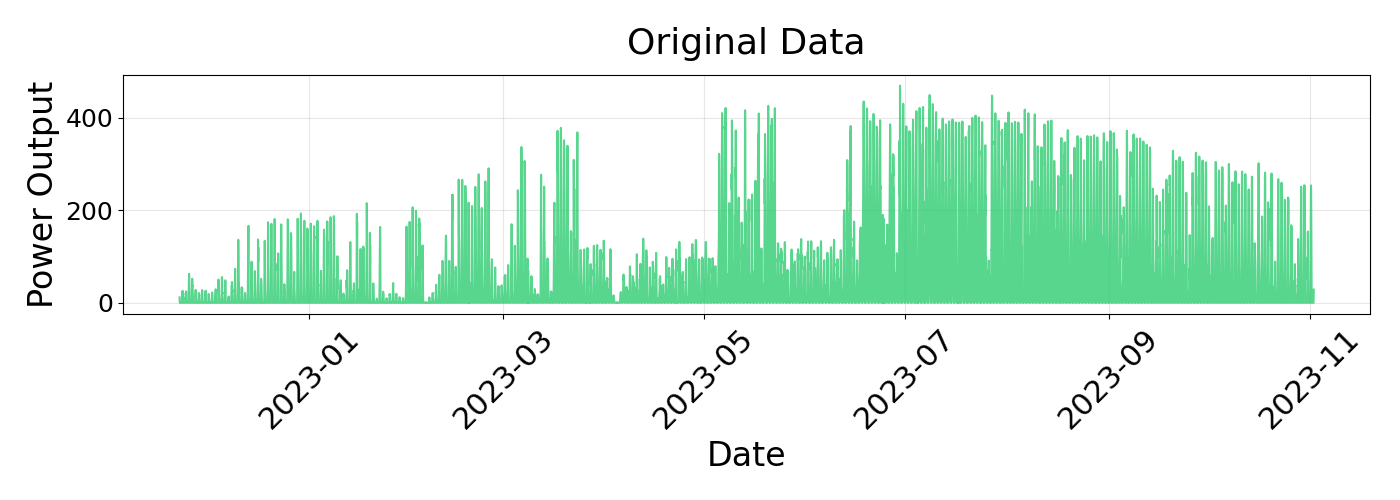}
        \includegraphics[width=1\linewidth]{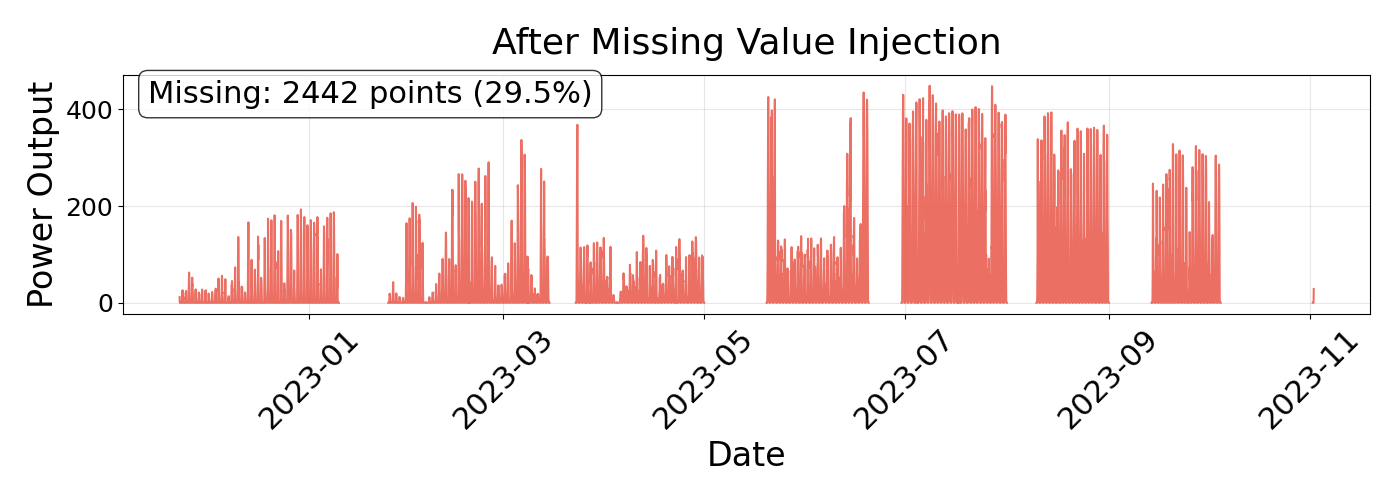} 
        \includegraphics[width=1\linewidth]{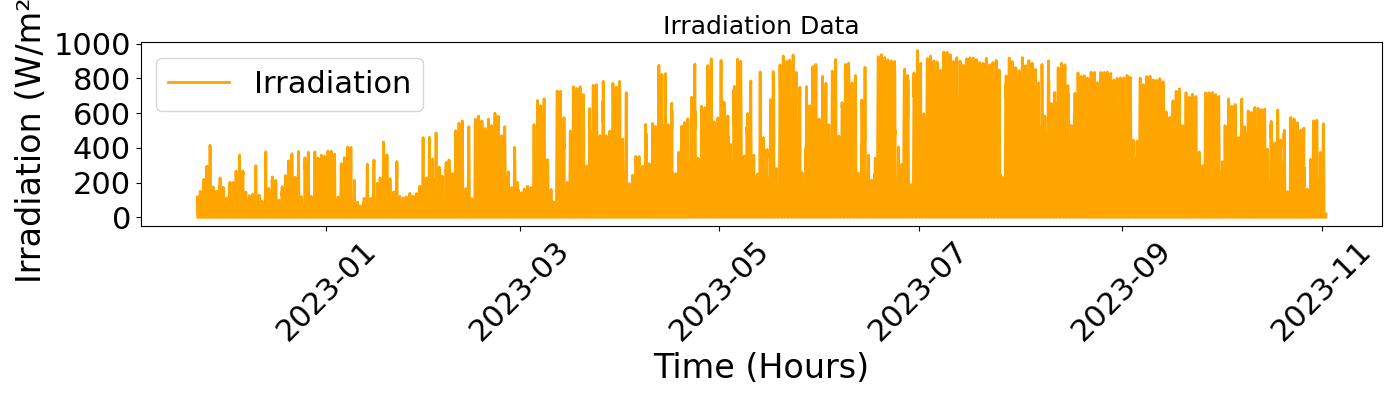}
    \caption{
    Dataset from the EU GRIDouble project used in our experiments.
    Top: original hourly DC power.
    Middle: DC power after block-wise removal of several contiguous weeks (29.5\% missing).
    Bottom: corresponding hourly irradiation (GHI).
    }
    \label{fig:missing_injection}
\end{figure}

The dataset used in this study was obtained from the European Union GRIDouble project\footnote{https://github.com/vodena/GRIDouble}. It contains hourly PV power generation and solar irradiation data collected from industrial facilities in Čačak, Serbia, covering November 2022 to November 2023.

The dataset includes approximately 8{,}275 hourly observations of generated energy (DC Power) across three locations, measured in kilowatt-hours (kWh) (Figure~\ref{fig:missing_injection}). For irradiation, we use the Global Horizontal Irradiance (GHI), measured in kilowatts per square meter (kW/m²). The training data consists of hourly records from 2022/11/22 to 2023/08/25, and the remaining 1{,}650 hours are reserved for testing. The split is chronological, and no shuffling is applied.

To evaluate the impact of missing values in a controlled manner, we simulate missingness by removing several contiguous weeks from both the training and test sets (Figure~\ref{fig:missing_injection}). This block-wise pattern reflects realistic device-level outages, such as inverter or sensor failures, which interrupt data recording for extended periods. In the resulting dataset, approximately 29.5\% of the PV power observations are missing. Although the experiments focus on block missingness, the proposed framework is not restricted to this pattern and can be applied to other missing mechanisms.

Throughout the experiments, irradiation is assumed to be fully observed, reflecting typical operational settings where meteorological measurements are reliably recorded.

Missing values are simulated rather than relying on naturally missing observations in order to retain access to the true underlying values. This enables objective evaluation of imputation accuracy and predictive coverage, which would not be possible if the ground truth were unavailable.

\subsection{Imputation Setups}

We consider three imputation settings in order to disentangle the impact of uncertainty arising from different stages of the pipeline: 
\begin{itemize}
    \item {\bf Setup 1}: Both the training data and the test input features are imputed using single imputation. 
    \item {\bf Setup 2}: The training data are imputed using single imputation, while the test input features are imputed using multiple imputation.  
    \item {\bf Setup 3}: Both the training data and the test input features are imputed using multiple imputation.
\end{itemize}
This design allows us to isolate the effect of uncertainty introduced at the training stage, the prediction stage, and their combination.

For single imputation, we use the kNNImputer~\cite{troyanskaya2001missing}, a widely used imputation method that replaces each missing value with the average of its nearest neighbors. For multiple imputation, we use kNNSampler~\cite{pashmchi2025knnsampler}, which generates stochastic imputations by sampling from the empirical conditional distribution induced by nearest neighbors (see Section~\ref{sec:imputation-knnsampler}).

The number of multiple imputations, $B$, is set to $B=5$ and $B=10$ for comparison.

\subsection{Machine Learning Models} 

We evaluate the proposed framework using four standard prediction models~\cite{hastie2009elements}:
Random Forest (RF)~\cite{breiman2001random},
k-Nearest Neighbors (kNN)~\cite{cover1967nearest},
a two-layer Multi-Layer Perceptron (MLP)~\cite{rumelhart1986learning},
and Lasso regression~\cite{tibshirani1996regression}.
All models are implemented in the Scikit-learn library~\cite{scikit-learn}.

Hyperparameters are tuned using time-series cross-validation via the 
\texttt{TimeSeriesSplit} procedure in Scikit-learn, which preserves chronological order by training on past observations and validating on future data. No shuffling is applied.

For kNN, the number of neighbors $k \in \{1,\dots,500\}$ is selected by 10-fold time-series cross-validation.  
RF is tuned using \texttt{HalvingRandomSearchCV} with an initial ensemble of 600 trees, exploring maximum depth, minimum samples per split and leaf, and feature subsampling.  
The MLP consists of two hidden layers (100 and 50 neurons) and is trained for 1000 iterations using the Adam optimizer, with hyperparameters selected via 5-fold time-series cross-validation.  
Lasso uses \texttt{LassoCV} to select the regularization parameter $\alpha$.

\subsection{Evaluation Metrics} {\label{sec:metrics}}

Predictive performance is assessed using two metrics: \\

\noindent
\textbf{Coverage Probability (CP).}
Let $[L_t, U_t]$ denote the $100(1-\alpha)\%$ prediction interval
for the test observation $Y_t^{\rm te}$ constructed in Section~\ref{sec:interval_estimation}, under either the normal or the gamma predictive distribution.

Coverage is evaluated only on test points defined in~\eqref{eq:test-missing} for which the true value is observed, i.e., $M_t^{\rm te} = 0$. Define
$
\mathcal{T}_{\rm obs}
=
\{ t : M_t^{\rm te} = 0 \}.
$
The empirical coverage probability is
\begin{equation*}
\mathrm{CP}
=
\frac{1}{|\mathcal{T}_{\rm obs}|}
\sum_{t \in \mathcal{T}_{\rm obs}}
\mathbf{1}
\left\{
Y_t^{\rm te} \in [L_t, U_t]
\right\}.
\end{equation*}
If $\mathrm{CP} < 1-\alpha$, the intervals are narrower than the designed level (overconfident). If $\mathrm{CP} > 1-\alpha$, they are wider than the designed level (conservative).  \\

\noindent
\textbf{Normalized Root Mean Square Error (NRMSE).}

Prediction accuracy is measured by the normalized root mean square error (NRMSE), computed over observed test points. Let $\widehat{Y}_t^\te$ denote the predictive mean in~\eqref{eq:pred-mean-final}. Then NRMSE is defined as
\begin{equation*}
\mathrm{NRMSE}
=
\sqrt{
\frac{1}{|\mathcal{T}_{\rm obs}|}
\sum_{t \in \mathcal{T}_{\rm obs}}
\left(
Y_t^{\rm te} - \widehat{Y}_t^\te
\right)^2
} ~/~ Y_{\max},
\end{equation*}
where $Y_{\max} = \max_{t \in \mathcal{T}_{\rm obs}} Y_t^{\rm te}$.

\section{Results} {\label{sec:results}}

\begin{table}[h]
\centering
\renewcommand{\arraystretch}{1.55}
\setlength{\tabcolsep}{7pt}
\caption{Coverage probability (\%) of 95\% prediction intervals under the normal distribution}
\label{table:coverage-normal}

\begin{tabular}{l|c|c|c|c|c}
\hline
\makecell{Imputation Setup} & $B$ & RF & kNN & MLP & Lasso \\
\hline

1) SI train, SI test
& -- & 74.4 & 83.8 & 83.0 & 85.1 \\
\hline

2) SI train, MI test
& 5  & 84.1 & 88.5 & 90.0 & 90.7 \\
& 10 & 83.9 & 88.1 & 90.5 & 91.5 \\
\hline

3) MI train, MI test
& 5  & 85.2 & 92.1 & 93.1 & 93.4 \\
& 10 & 84.5 & 92.1 & 93.6 & 92.8 \\
\hline

\end{tabular}
\end{table}

\begin{table}[h]
\centering
\renewcommand{\arraystretch}{1.55}
\setlength{\tabcolsep}{7pt}
\caption{Coverage probability (\%) of 95\% prediction intervals under the gamma distribution}
\label{table:coverage-gamma}

\begin{tabular}{l|c|c|c|c|c}
\hline
Imputation Setup & $B$ & RF & kNN & MLP & Lasso \\
\hline

1) SI train, SI test 
& -- & 70.7 & 78.8 & 72.1 & 70.1 \\
\hline

2) SI train, MI test
& 5  & 80.7 & 83.1 & 78.7 & 77.0 \\
& 10 & 80.9 & 83.6 & 79.5 & 76.5 \\
\hline

3) MI train, MI test 
& 5  & 83.7 & 88.1 & 78.3 & 79.5 \\
& 10 & 82.9 & 88.7 & 84.7 & 79.2 \\
\hline

\end{tabular}
\end{table}

\begin{table}[h]
\centering
\renewcommand{\arraystretch}{1.65}
\setlength{\tabcolsep}{8pt}
\caption{NRMSE for different algorithms under each imputation setup}
\label{table:benchmark-nrmse}

\begin{tabular}{l|c|c|c|c|c}
\hline
Imputation Setup & $B$ & RF & kNN & MLP & Lasso \\
\hline

1) SI train, SI test
& -- & 0.123 & 0.131 & 0.124 & 0.132 \\
\hline

2) SI train, MI test
& 5  & 0.127 & 0.136 & 0.126 & 0.132 \\
& 10 & 0.126 & 0.132 & 0.123 & 0.129 \\
\hline

3) MI train, MI test
& 5  & 0.127 & 0.136 & 0.125 & 0.132 \\
& 10 & 0.126 & 0.133 & 0.121 & 0.130 \\
\hline

\end{tabular}
\end{table}

\begin{figure}[htbp]
    \centering
    
    \begin{subfigure}[b]{\linewidth}
        \centering
        \includegraphics[width=0.80\linewidth]{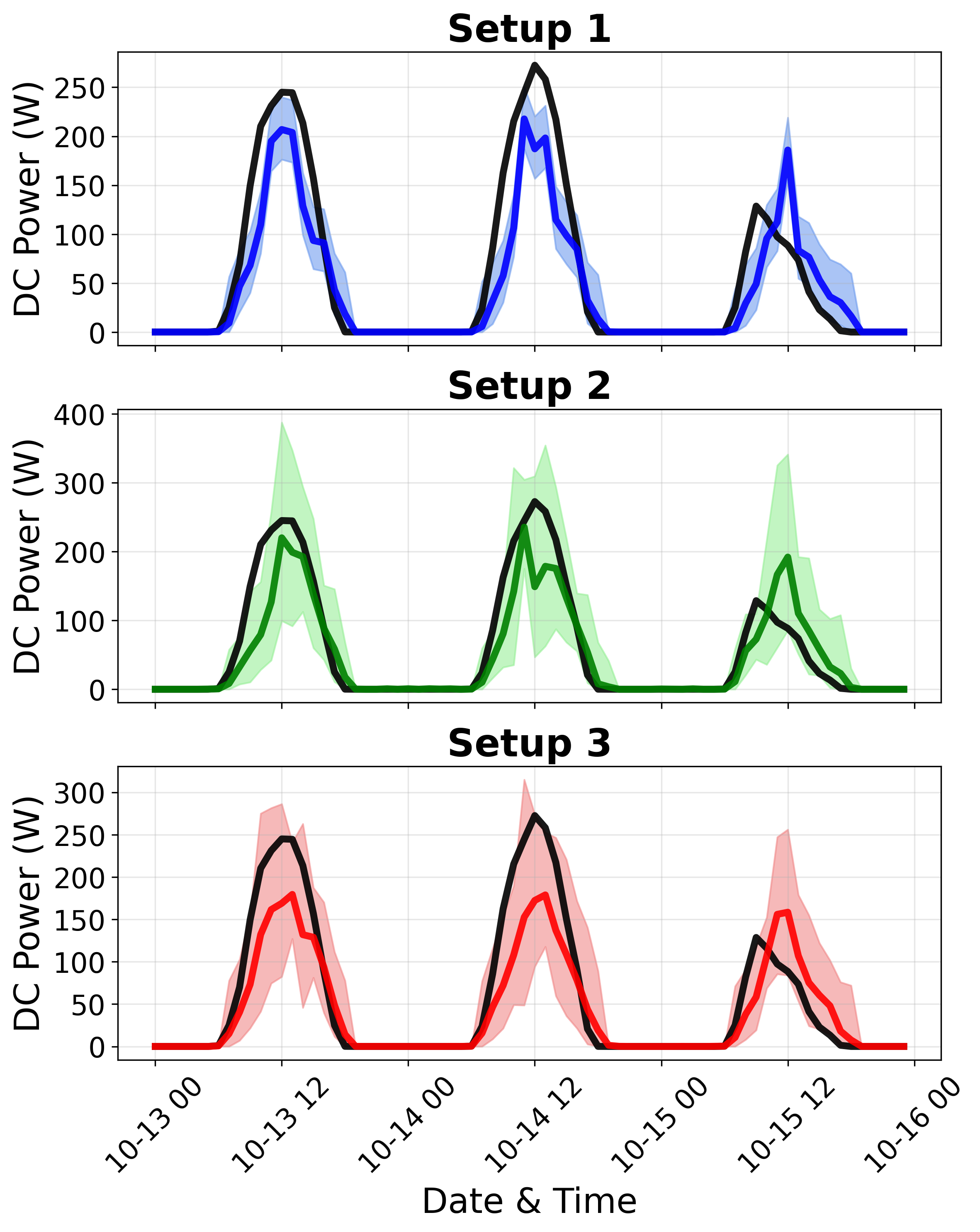}
        \caption{Gamma intervals with Random Forest}
        \label{fig:rf-gamma}
    \end{subfigure}
    
    \vspace{0.5em}
    
    \begin{subfigure}[b]{\linewidth}
        \centering
        \includegraphics[width=0.80\linewidth]{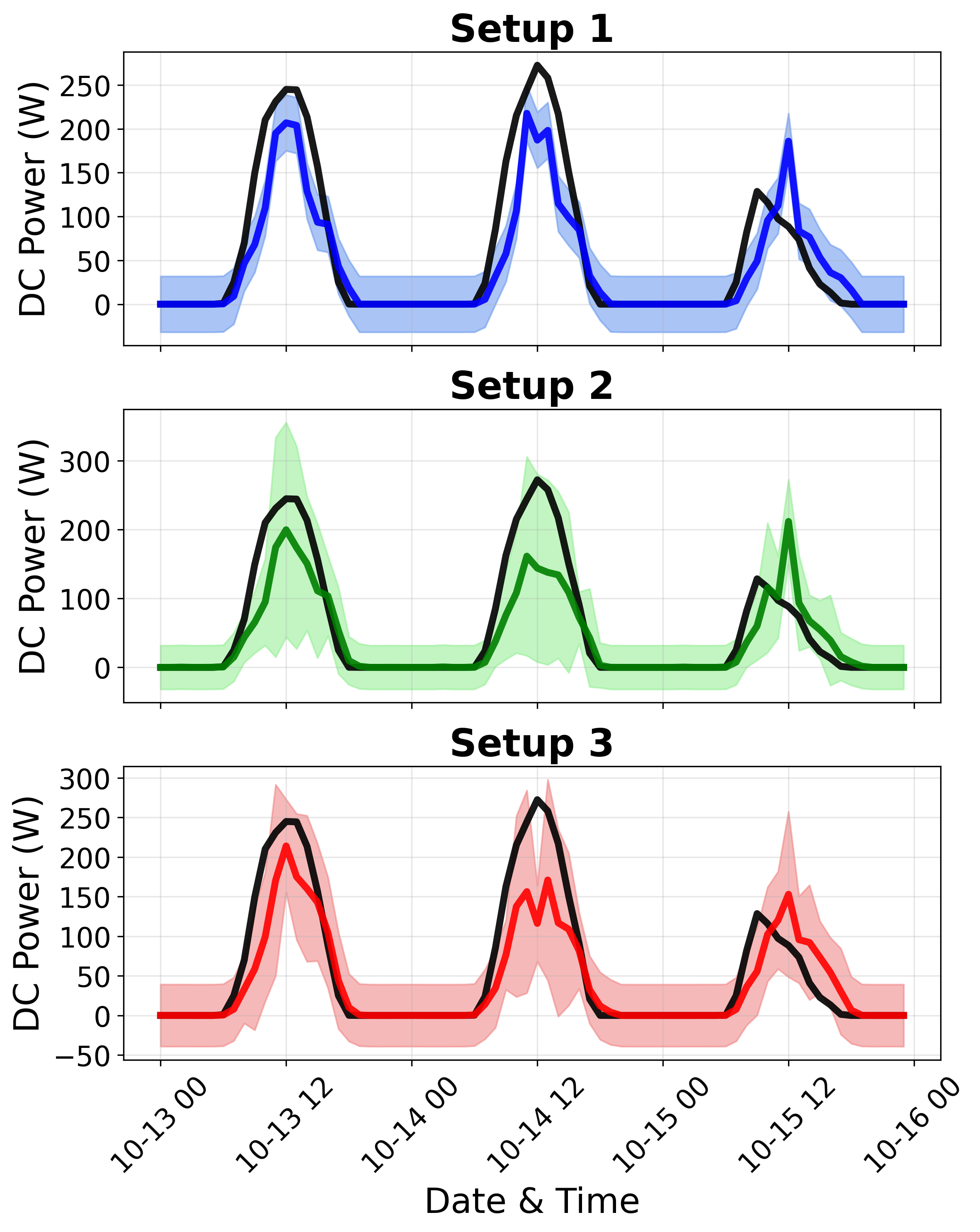}
        \caption{Normal intervals with Random Forest}
        \label{fig:rf-normal}
    \end{subfigure}

\caption{
95\% prediction intervals ($B=5$) for {\bf Random Forest} under (a) gamma and (b) normal predictive distributions, shown for three imputation setups: (1) SI train \& test, (2) SI train with MI test, and (3) MI train \& test. Black thick curve: ground truth; thick colored curve: predictive means; shaded region: 95\% prediction intervals.
}
    \label{fig:PI-rf-comparison}
\end{figure}

\begin{figure}[htbp]
    \centering
    
    \begin{subfigure}[b]{\linewidth}
        \centering
        \includegraphics[width=0.80\linewidth]{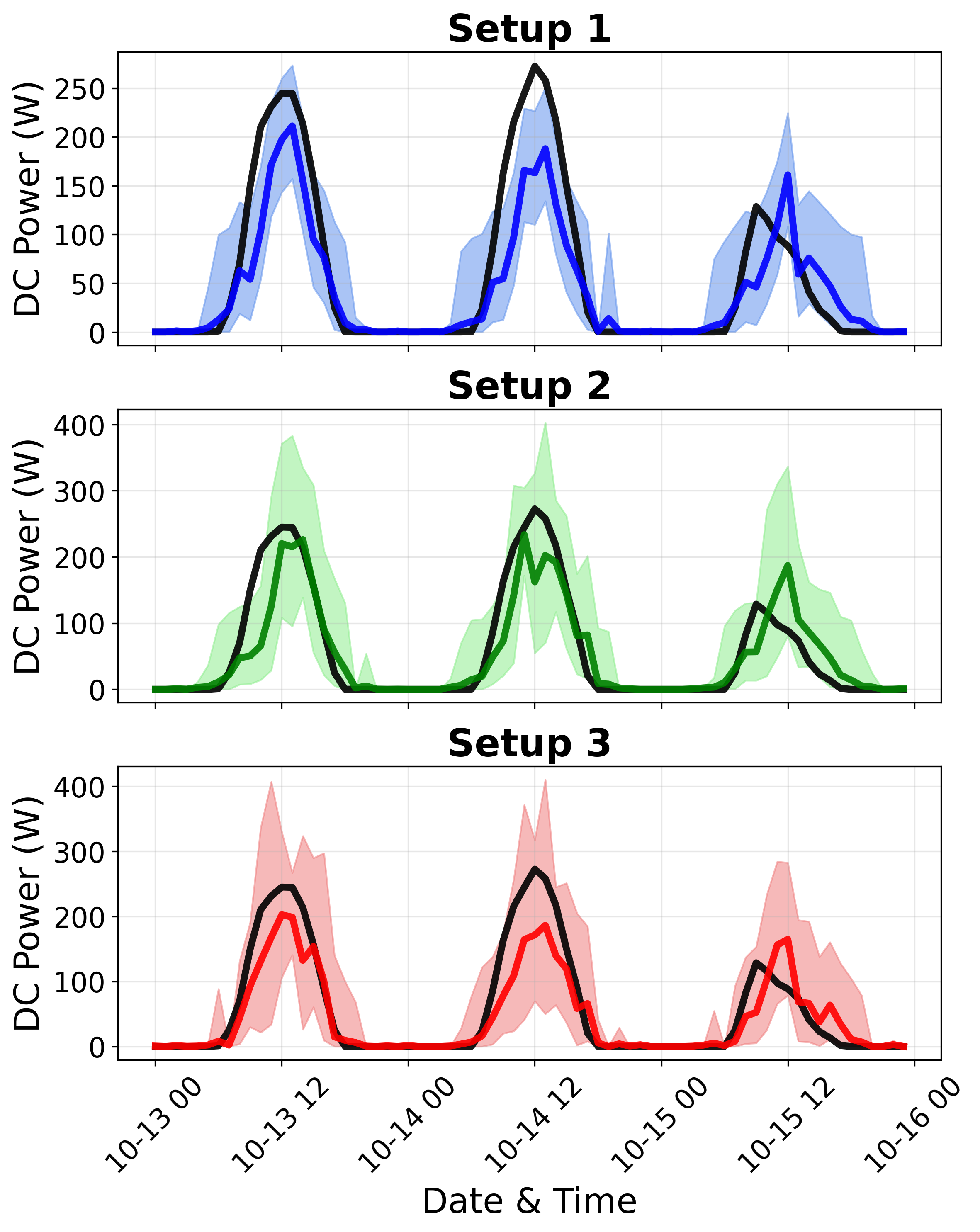}
        \caption{Gamma intervals with MLP}
        \label{fig:mlp-gamma}
    \end{subfigure}
    
    \vspace{0.5em}
    
    \begin{subfigure}[b]{\linewidth}
        \centering
        \includegraphics[width=0.80\linewidth]{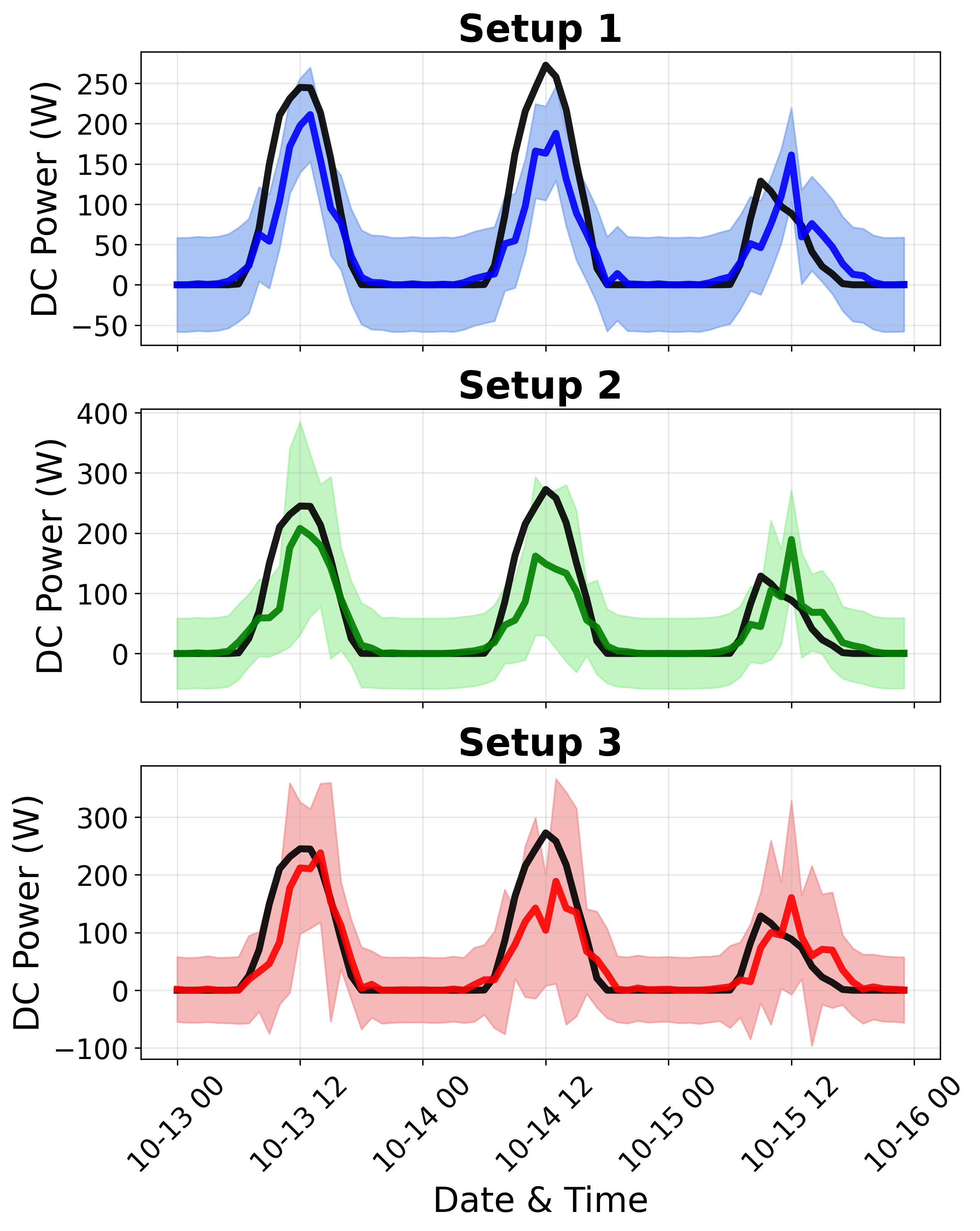}
        \caption{Normal intervals with MLP}
        \label{fig:mlp-normal}
    \end{subfigure}

\caption{
95\% prediction intervals ($B=5$) for {\bf MLP} under (a) gamma and (b) normal predictive distributions, shown for three imputation setups: (1) SI train \& test, (2) SI train with MI test, and (3) MI train \& test. Black thick curve: ground truth; thick colored curve: predictive means; shaded region: 95\% prediction intervals.
}
    \label{fig:PI-mlp-comparison}
\end{figure}

\begin{figure}[htbp]
    \centering
    
    \begin{subfigure}[b]{\linewidth}
        \centering
        \includegraphics[width=0.80\linewidth]{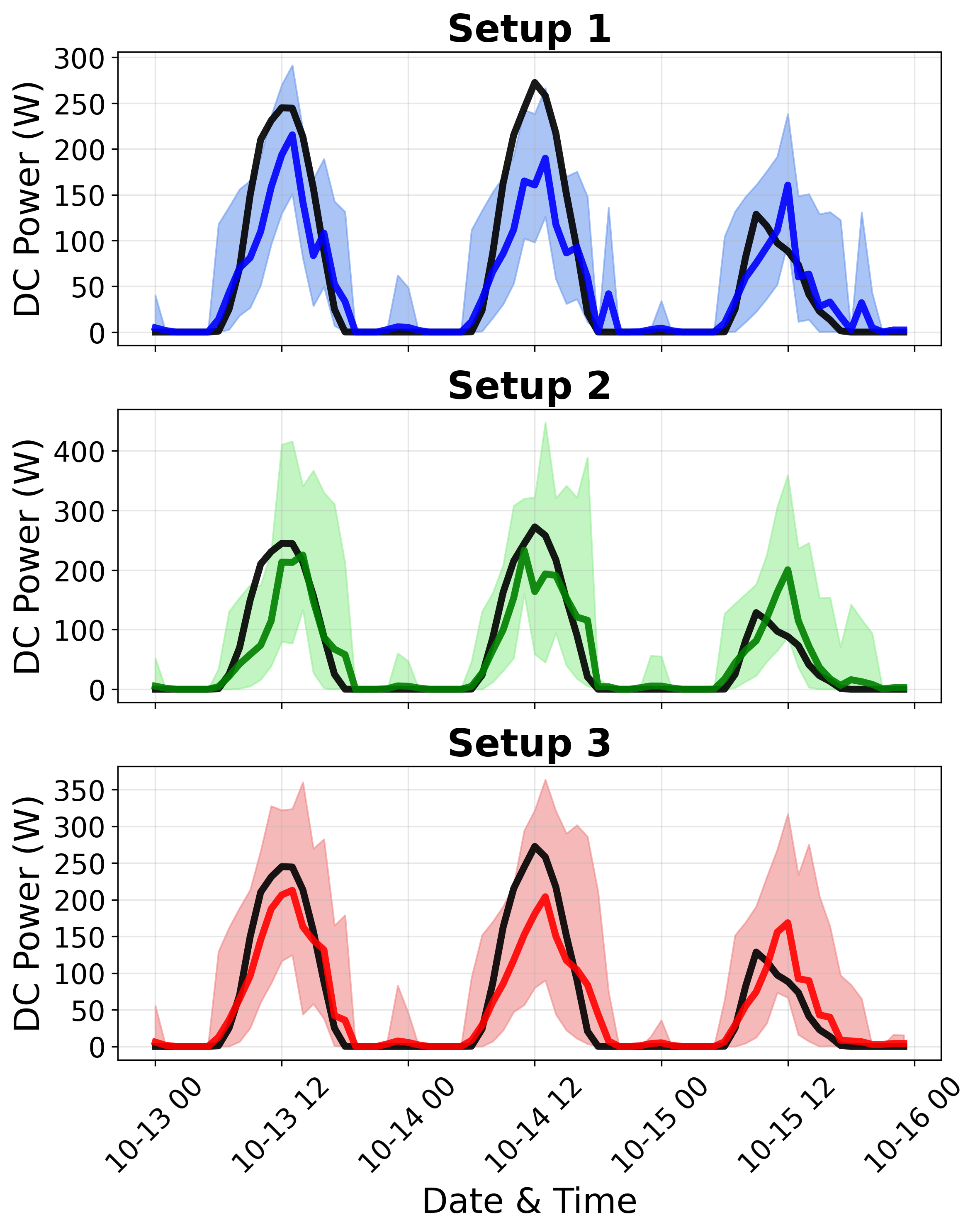}
        \caption{Gamma intervals with Lasso}
        \label{fig:lasso-gamma}
    \end{subfigure}
    
    \vspace{0.5em}
    
    \begin{subfigure}[b]{\linewidth}
        \centering
        \includegraphics[width=0.80\linewidth]{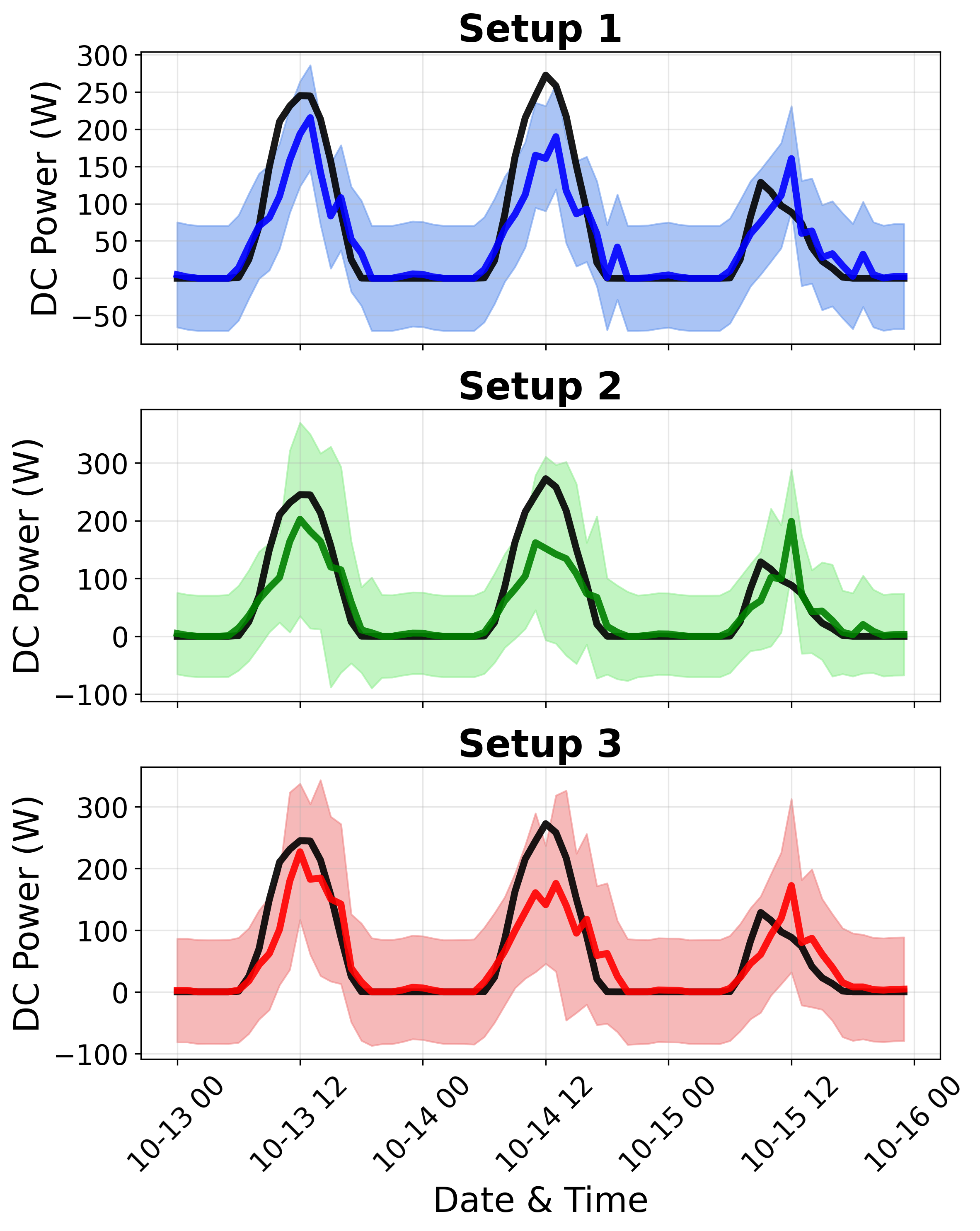}
        \caption{Normal intervals with Lasso}
        \label{fig:lasso-normal}
    \end{subfigure}

\caption{
95\% prediction intervals ($B=5$) for {\bf Lasso} under (a) gamma and (b) normal predictive distributions, shown for three imputation setups: (1) SI train \& test, (2) SI train with MI test, and (3) MI train \& test. Black thick curve: ground truth; thick colored curve: predictive means; shaded region: 95\% prediction intervals.
}
    \label{fig:PI-lasso-comparison}
\end{figure}

\begin{figure}[htbp]
    \centering
    
    \begin{subfigure}[b]{\linewidth}
        \centering
        \includegraphics[width=0.80\linewidth]{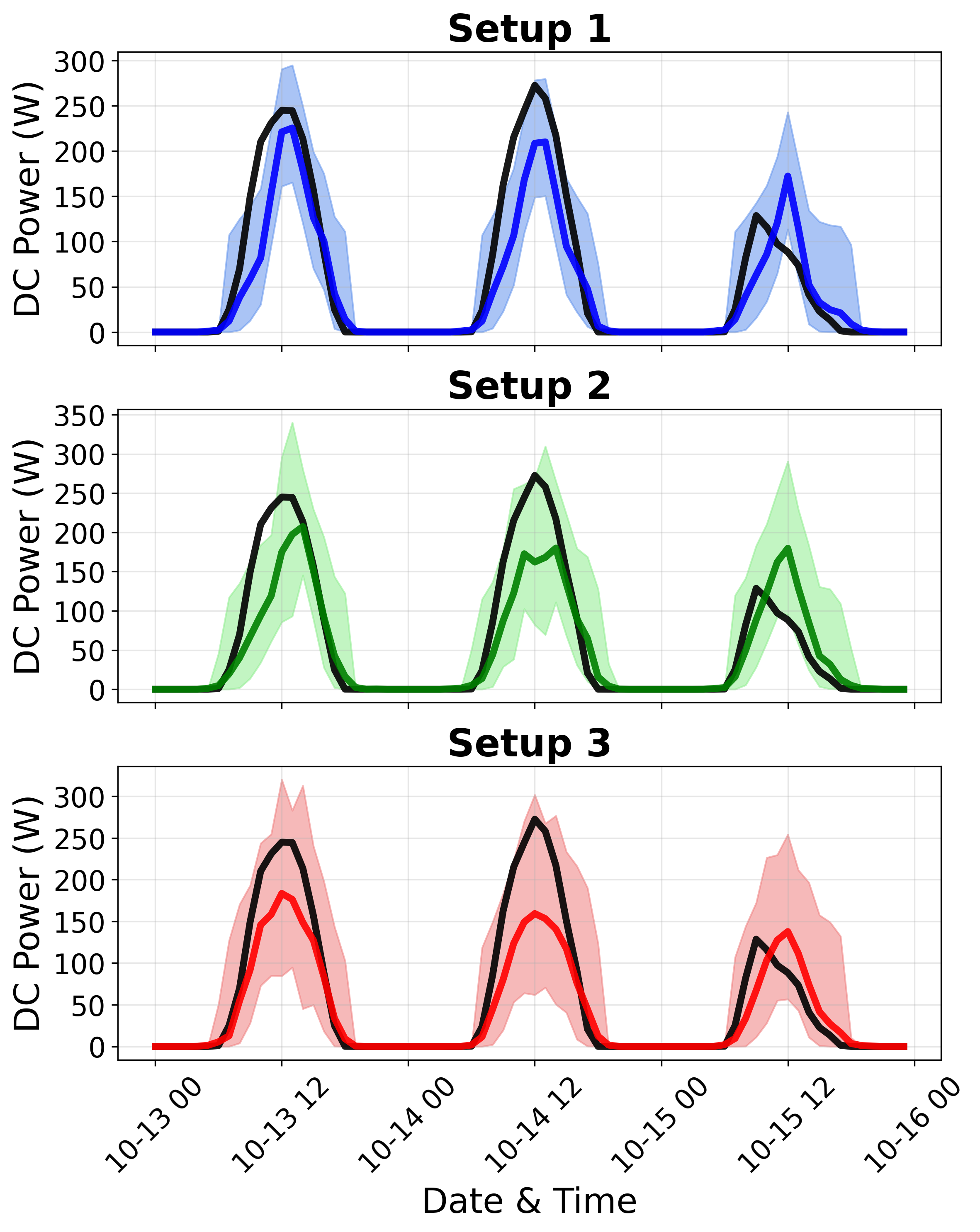}
        \caption{Gamma intervals with kNN}
        \label{fig:knn-gamma}
    \end{subfigure}
    
    \vspace{0.5em}
    
    \begin{subfigure}[b]{\linewidth}
        \centering
        \includegraphics[width=0.80\linewidth]{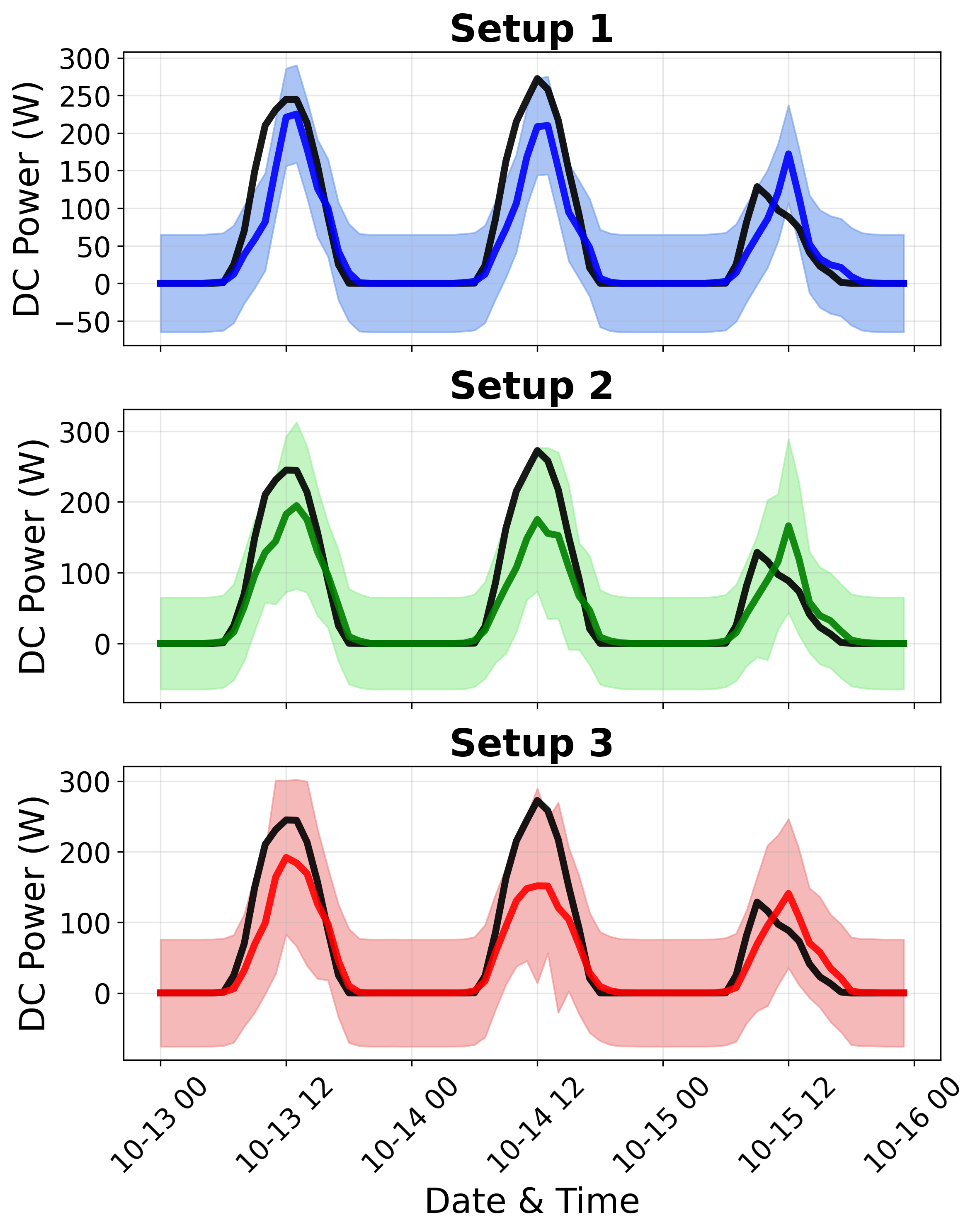}
        \caption{Normal intervals with kNN}
        \label{fig:knn-normal}
    \end{subfigure}

\caption{
95\% prediction intervals ($B=5$) for {\bf kNN} under (a) gamma and (b) normal predictive distributions, shown for three imputation setups: (1) SI train \& test, (2) SI train with MI test, and (3) MI train \& test. Black thick curve: ground truth; thick colored curve: predictive means; shaded region: 95\% prediction intervals.
}
    \label{fig:PI-knn-comparison}
\end{figure}

Tables~\ref{table:coverage-normal} and \ref{table:coverage-gamma} report the empirical coverage probabilities of the 95\% prediction intervals under the normal and gamma predictive distributions, respectively. 
Table~\ref{table:benchmark-nrmse} reports the corresponding NRMSE values. 
Figures~\ref{fig:PI-rf-comparison}–\ref{fig:PI-knn-comparison} describe prediction intervals over arbitrarily selected three days for all predictors, illustrating that the effect of multiple imputation is not model-specific. 
Throughout this section, SI and MI denote single and multiple imputation, respectively.

The following observations can be made from the results: \\

\noindent
{\bf MI corrects overconfidence under SI.}
For each predictor and each predictive distribution, Setup~2 (SI train, MI test) and Setup~3 (MI train, MI test) yield higher coverage than Setup~1 (SI train, SI test). 
This indicates that multiple imputation corrects the overconfidence induced by single imputation, which treats missing values as fixed. 
For example, for the MLP with normal intervals, the coverage increases from 83.0\% under Setup~1 to 90.0\% under Setup~2 ($B=5$) and 93.1\% under Setup~3 ($B=5$), the latter being close to the nominal 95\% level.\\

\noindent
{\bf Uncertainty in test inputs dominates that in training data.}
The increase in coverage from Setup~1 (SI train, SI test) to Setup~2 (SI train, MI test) is larger than the change from Setup~2 to Setup~3 (MI train, MI test). 
For example, for Random Forest with normal intervals ($B=5$), coverage increases by 9.7 percentage points from Setup~1 to Setup~2 (74.4\% to 84.1\%), whereas the increase from Setup~2 to Setup~3 is only 1.1 points (84.1\% to 85.2\%). 
This suggests that most of the additional predictive uncertainty arises from missing values in the test input features rather than from the training stage. \\

\noindent
{\bf A small number $B$ of imputations suffices.}
The coverage probabilities and NRMSE values differ little between $B=5$ and $B=10$ for most cases. 
This suggests that a small number of imputations, such as $B=5$, is adequate in practice. 
This is consistent with Rubin's observation that “as few as five (or even three in some cases) is adequate” \cite[p.~480]{Rubin96}.\\

\noindent
{\bf MI does not materially affect prediction accuracy.}
For each predictor, the NRMSE values are similar across the three imputation setups. 
Thus, the improvement in coverage under multiple imputation is not accompanied by a material change in predictive accuracy. \\

\noindent
{\bf Normal intervals achieve better calibration.}
Gamma intervals respect the non-negativity of PV power and
serve as a simple non-negative baseline. In this study, however,
their empirical coverages are substantially below the nominal
95\% level, even under Setup 3.
In contrast, normal intervals under Setup~3 are close to the nominal level for kNN, MLP, and Lasso. 
Thus, within this study, the normal formulation yields better calibrated prediction intervals. 
If strict non-negativity is required, negative lower bounds can be truncated in practice, since the observed PV output is non-negative. \\

\noindent
{\bf Normal intervals with MI for both training and test data achieve the best calibration.}
In summary, normal prediction intervals under Setup~3 (MI train, MI test), corresponding to the proposed imputation framework described in Section~\ref{sec:multiple-imput-frame}, achieve the best calibration among the considered settings.

\section{Conclusion}

This study introduced a principled framework for propagating missing-data uncertainty into predictive distributions for short-term PV forecasting.

Existing PV forecasting studies treat imputed values as fixed and do not propagate imputation uncertainty into predictive distributions. In contrast, we integrate stochastic multiple imputation with Rubin’s rule to quantify and propagate the additional variance induced by missing observations.

Empirical results show that ignoring this source of uncertainty leads to systematically under-covered intervals. Accounting for it restores calibration without degrading point prediction accuracy.

The experiments were conducted on one real dataset with simulated block missingness, which allows objective evaluation because the ground truth is retained. Broader validation across sites, climates, and missingness mechanisms remains for future work.

Overall, missing data is not merely a preprocessing issue. It is a source of predictive uncertainty that must be modeled explicitly when probabilistic forecasts are used for operational decision-making.


\bibliographystyle{IEEEtran}
\bibliography{bibfile-ieee}
\vspace{-4\baselineskip}
\begin{IEEEbiography}[{\includegraphics[width=1in,height=1.25in,clip,keepaspectratio]{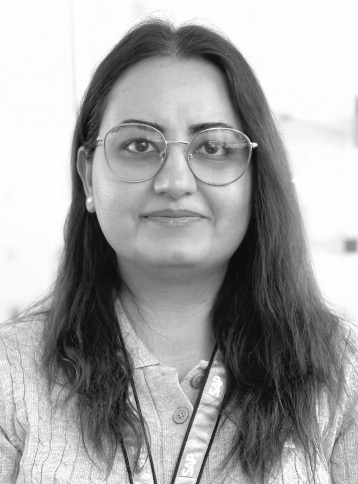}}]{\textbf{Parastoo Pashmchi}} received the Ph.D. degree in data science from Sorbonne University, in 2026, through a joint industrial program with SAP Labs France. She received the M.Sc. degree in smart cities engineering from Université Côte d’Azur, Nice, France, in 2022. Previously, she earned another M.Sc. degree in civil engineering from Isfahan University of Technology, Isfahan, Iran, in 2018. She received her B.Sc. degree in civil engineering in Iran in 2014. Her research focuses on smart infrastructures and energy transition systems.
\end{IEEEbiography}
\vspace{-4\baselineskip}
\begin{IEEEbiographynophoto}{J\'er\^ome Benoit}
received the Engineer's degree in computer science from Polytech'Marseille, Marseille, France, in 2019. He previously received the M.Sc.\ degree in mathematics from Aix-Marseille Université, Marseille, France, in 1997. He has been working as an R\&D Software Engineer at SAP Labs France since 2019. His research focuses on e-mobility, electric vehicle charging infrastructure, and machine learning applied to smart charging systems.
\end{IEEEbiographynophoto}

\vspace{-4\baselineskip}
\begin{IEEEbiography}[{\includegraphics[width=1in,height=1.25in,clip,keepaspectratio]{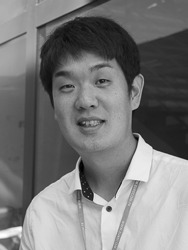}}]%
{Motonobu Kanagawa} is an Assistant Professor in the Data Science Department at EURECOM, France, since 2019.  
Previously, he was a research scientist at the Max Planck Institute for Intelligent Systems and the University of Tuebingen in Germany from 2017 to 2019. Before this, he was a postdoctoral researcher (2016-2017) and a PhD student (2013-2016) at the Institute of Statistical Mathematics in Japan.  
\end{IEEEbiography}

\end{document}